\documentclass[11pt]{article}

\usepackage[final]{acl}

\usepackage{times}
\usepackage{latexsym}
\usepackage{ragged2e}

\usepackage[T1]{fontenc}

\usepackage{microtype}

\usepackage{inconsolata}
\usepackage{booktabs}
\usepackage{graphicx}

\usepackage{inconsolata}
\usepackage{listings}
\usepackage{xcolor}  

\usepackage{booktabs}     
\usepackage{caption}
\usepackage{subcaption}   

\usepackage[english]{babel}
\usepackage{amsmath}   
\usepackage{amssymb}   
\usepackage{algorithm}
\usepackage{algpseudocode}
\usepackage{amsmath}
\usepackage{authblk}
\usepackage{stfloats}
\usepackage{float}

\usepackage{amsmath}
\usepackage{graphicx}
\title{Sign Language Video Synthesis via Loss-Guided Multi-Expert GANs (Technical Report)
\large }

\author[1]{Nong Dingzhan}
\author[1]{REN Zhihao}
\author[1]{Li Ziqi}
\author[1]{Tim Lo\thanks{Corresponding author: Tim Lo, \texttt{admin@xai.hk}}}

\affil[1]{Glassbox AI}
\affil[ ]{\texttt{nongdingzhan@gmail.com}, \texttt{renzhihao0919@gmail.com}, \texttt{ziq93812@gmail.com}, \texttt{tim.lo@xai.hk}}

\begin{document}
\maketitle

\begin{abstract}
This preliminary technical report presents a framework for sign language video synthesis using a loss-guided multi-expert Generative Adversarial Network (GAN) to enhance communication for individuals with hearing impairments. Three specialized discriminators---global, hand, and head---each guide a corresponding expert branch in the generator toward a distinct visual region, enabling implicit feature specialization without explicit diversity losses. To stabilize this multi-discriminator system, whose early-phase training otherwise exhibits chaotic dynamics, we introduce a United Loss consensus mechanism that regularizes each discriminator toward the ensemble average at a 10\% weight. Each branch further adopts a dual-pathway convolutional-transformer design with learnable AdaptiveFeatureFusion, balancing the stability of convolutions against the detail of windowed self-attention. The generator is trained using an alternating three-mode schedule (discriminator, holistic generation, branch-specialized generation). On a custom 156GB dataset with a filtered evaluation set that removes easy and repetitive samples, our 0.2B-parameter variant achieves 29.78 PSNR (0.9593 SSIM), the 0.66B variant reaches 30.52 PSNR (0.9631 SSIM) after 7.37M steps, and the 1.3B variant achieves 30.72 PSNR (0.9650 SSIM). The gain from 0.66B to 1.3B is only +0.20 PSNR despite nearly doubling the parameters, demonstrating sharply diminishing returns. Inference VRAM footprints are 1.5 GB, ~5 GB, and 8 GB respectively, enabling deployment on consumer-grade hardware. Full ablation studies remain ongoing due to the 2--3 month training cycle on a single GPU. The system was showcased at the 2025 Hong Kong Frontier Technology Summit.
\end{abstract}

\section{Introduction}
Sign language video synthesis is critical for improving communication accessibility for individuals with hearing impairments, requiring high-fidelity video generation with precise hand gestures and facial expressions. While Generative Adversarial Networks (GANs) have advanced image and video synthesis, traditional single-discriminator architectures struggle to capture fine-grained details in complex scenes like sign language videos. This limitation often results in unrealistic outputs, particularly for intricate hand and face movements.

To address these challenges, we propose a Multi-Discriminator GAN (MD-GAN) framework with specialized discriminators for global coherence, hand gestures, and facial expressions. This approach ensures detailed supervision of region-specific features. Additionally, we introduce a Multi parallel U-Net generator, where three parallel encoder-decoder branches focus on distinct visual regions (full body, hands, and face), with each branch driven by its corresponding discriminator to encourage implicit feature specialization.

The use of multiple discriminators can introduce training instability due to conflicting optimization objectives. To mitigate this, we propose a United Loss function that establishes a soft consensus constraint across discriminators. Each discriminator's loss is blended with the ensemble average at a 10\% weight, penalizing extreme deviation from the group while preserving individual specialization. This mechanism promotes stable convergence without sacrificing the benefit of region-specific feedback.

To ensure realistic outputs, we incorporate Adaptive Instance Normalization (AdaIN) to fuse skeleton structure with style image appearance at each encoder layer. Our model is designed for consumer-grade hardware: all variants train on a single GPU and infer within 1.5--8\,GB VRAM.

Our contributions include: (1) a loss-guided multi-expert framework in which specialized discriminators for hand, face, and global regions each drive a corresponding generator branch; (2) a Multi parallel U-Net framework with a dual-pathway architecture---each branch independently processes features through parallel convolutional and transformer pathways fused via learnable weights (AdaptiveFeatureFusion), enabling stable, region-specific feature extraction; (3) a United Loss function for federated discriminator consensus; and (4) an alternating three-mode training schedule that decouples holistic generation from branch-specialized optimization. We report PSNR and SSIM results across three model scales on a filtered evaluation set that explicitly targets challenging cases.

The paper is organized as follows: Section~2 reviews related work, Section~3 details the proposed methodology, Section~4 presents experimental results, and Section~5 concludes 
with limitations and future directions.

\section{Related Work}

\subsection{Generative Adversarial Networks for Video Synthesis} 
Generative Adversarial Networks (GANs) \cite{goodfellow2014generativeadversarialnetworks} have revolutionized the field of generative modeling. GANs consist of two neural networks---the Generator (G) and the Discriminator (D)---trained adversarially, where G generates synthetic data while D distinguishes between real and fake samples. This framework has inspired numerous advancements in deep learning, leading to improved architectures, training stability, and applications across various domains. A series of updates related to GAN have been introduced, like DCGAN~\cite{radford2016unsupervisedrepresentationlearningdeep}, WGAN~\cite{arjovsky2017wassersteingan}, Cycle GAN~\cite{zhu2020unpairedimagetoimagetranslationusing}, and especially, the Style GAN series~\cite{karras2019stylebasedgeneratorarchitecturegenerative}~\cite{Karras2019stylegan2}~\cite{Karras2021}.

\subsection{Style GAN}
The Style GAN series advances image synthesis in GANs. Style GAN~\cite{karras2019stylebasedgeneratorarchitecturegenerative} introduces a generator inspired by style transfer, separating high-level attributes from stochastic variations with scale-specific control. Style GAN2~\cite{Karras2019stylegan2} improves image quality by fixing artifacts through redesigned normalization and regularization. Style GAN3~\cite{Karras2021} ensures translation and rotation equivariance by addressing aliasing, suiting video applications.

\subsection{Multi Parallel U-Net}
In the realm of advanced image processing, Zhang et al.~\cite{ALJOWAIR2023104767} propose the Multi Parallel U-Net, a novel architecture that enhances the Mixture-of-Experts (MoE) paradigm for computer vision tasks. Unlike conventional MoE frameworks that rely on dynamic expert selection, the Multi Parallel U-Net activates all expert branches simultaneously for every input, ensuring uniform feature extraction and balanced gradient flow. While our architecture draws inspiration from this fully-parallel paradigm, it differs in two key respects: (1)~our branches share identical input resolution rather than employing pre-trained heterogeneous encoders, and (2)~our design introduces a United Loss consensus mechanism to coordinate training across branches, which is not present in the original Multi Parallel U-Net. Our v4 architecture further departs from the standard Multi Parallel U-Net by adopting a dual-pathway design within each branch, where parallel convolutional and Swin Transformer streams~\cite{liu2021Swin} are fused via learnable weights rather than fixed aggregation, enabling dynamic, per-branch specialization.

\subsection{Multi Discriminators}
In traditional GAN training, the generator tends to overly focus on partial modes of the real data distribution (e.g., repeatedly generating similar samples) while neglecting diversity, resulting in outputs that lack the high complexity of real data. However, existing classic solutions face limitations: methods like GMAN and D2GAN train discriminators independently without collaborative specialization, often yielding redundant feedback.

By contrast, in MCL-GAN~\cite{choi2024mclgangenerativeadversarialnetworks}, each discriminator automatically concentrates on specific subsets of the dataset (e.g., distinct facial poses or texture regions in CelebA) through the MCL loss function, establishing an ``expert specialization'' pattern. This approach demonstrates robust generalization capabilities even in low-sample scenarios.

Our task involves sign language generation, where the data distribution complexity for facial and hand details significantly exceeds that of other components. While a multi-discriminator architecture aligns with these inherent requirements, our dataset features locally labeled guidance. Thus, we pioneer a novel training methodology for multi-discriminators within this framework.

\section{Methodology}

\subsection{Multi-Discriminator}
The Multi-Discriminator GAN (MD-GAN) comprises three discriminators: a Global Discriminator for overall video coherence, a Hand Discriminator for gesture details, and a Head Discriminator for facial expressions, ensuring precise supervision of sign language video synthesis.

\paragraph{Global Discriminator}
The Global Discriminator processes 448$\times$448 images across five resolution levels (448, 224, 112, 56, 28), using Haar wavelet transforms to decompose inputs into frequency subbands (6 to 24 channels). It employs MiniBatch Standard Deviation (MBSD) to prevent mode collapse and outputs a 30$\times$30 feature map via a PatchGAN structure to enhance detail supervision.

\paragraph{Local Discriminators}
Local discriminators operate on 112$\times$112 patches across three scales (112, 56, 28). The Hand Discriminator extracts hand regions using keypoint-based localization, scaling to 112$\times$112 with center-crop alignment, focusing on finger positions and hand shapes. The Head Discriminator processes facial regions using nose keypoints with a 96-pixel radius, capturing expression details.

\paragraph{Haar Wavelet Transforms}
The discriminator employs Haar wavelet transforms to decompose RGB images into frequency subbands, expanding the input channels from 6 to 24 (6$\times$4 subbands) to simultaneously capture structural (low-frequency) and textural (high-frequency) features. Additionally, it integrates MiniBatch Standard Deviation (MBSD) to compute statistical variations across mini-batches, enhancing discriminative power and preventing mode collapse by providing batch-level contextual information for improved real/fake sample distinction.

\subsection{Generator}

\begin{figure*}[tp]
    \centering
    \includegraphics[width=1\textwidth]{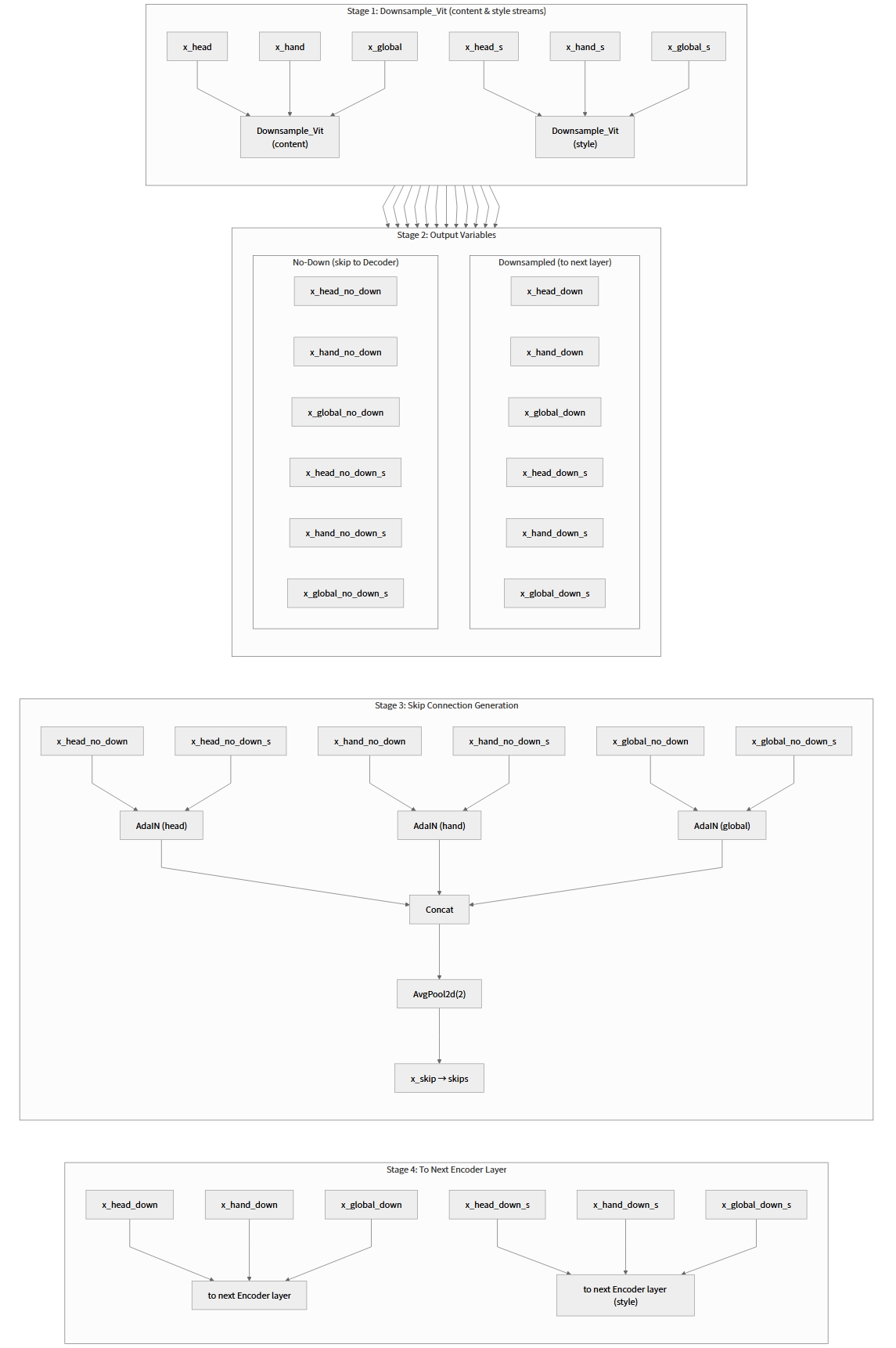}
    \caption{Encoder architecture with three parallel branches (head, hand, global). Each branch processes identical input through independent dual-pathway Downsample\_Vit blocks. AdaIN fuses content and style streams at each level to produce skip connections.}
    \label{fig:img_encoder}
\end{figure*}
\begin{figure}[t]
    \centering
    \includegraphics[width=0.5\textwidth]{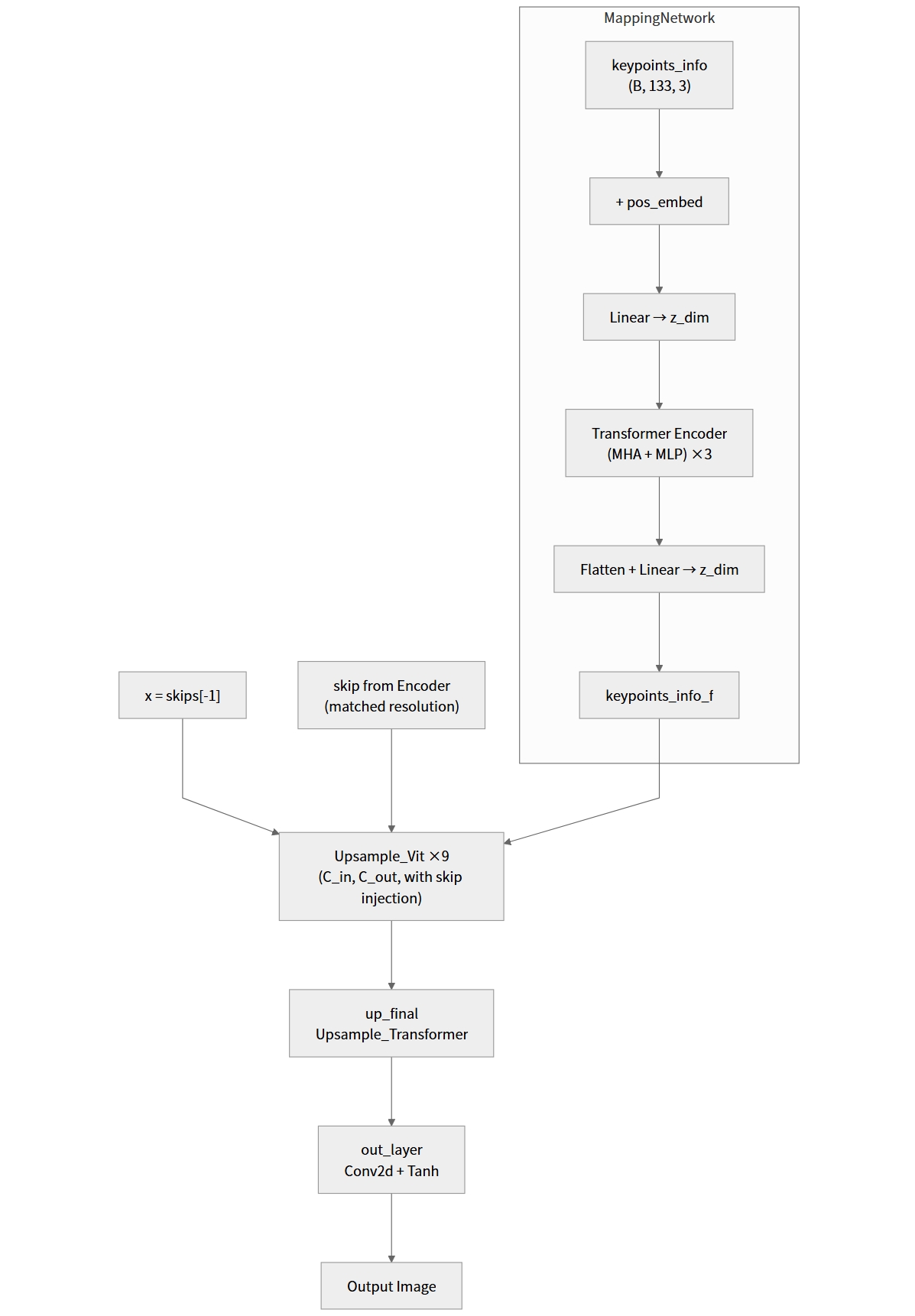}
    \caption{Decoder architecture. Three parallel encoder-decoder branches produce skip connections, concatenated and processed by Upsample\_Vit blocks with keypoint-guided cross-attention.}
    \label{fig:img_decoder}
\end{figure}

\paragraph{Multi-Branch Parallel Processing Architecture}
Multi parallel U-Net builds upon the Mixture-of-Experts (MoE) paradigm by introducing a fully parallel architecture that addresses expert load imbalance and training instability inherent in conventional MoE models. Unlike dynamic expert selection in MoE, Multi parallel U-Net simultaneously activates all branches for every input, ensuring uniform feature extraction coverage and balanced gradient flow. Moreover, each branch is driven by its corresponding specialized discriminator (hand, face, or global), which provides implicit pressure for the branch to specialize on specific visual regions without requiring explicit diversity losses. Adaptive Instance Normalization (AdaIN) is introduced to implement the image translation task in the form of style transfer. The overall encoder structure is illustrated in Figure~\ref{fig:img_encoder}.

\begin{figure*}[tp]
    \centering
    \includegraphics[width=0.8\textwidth]{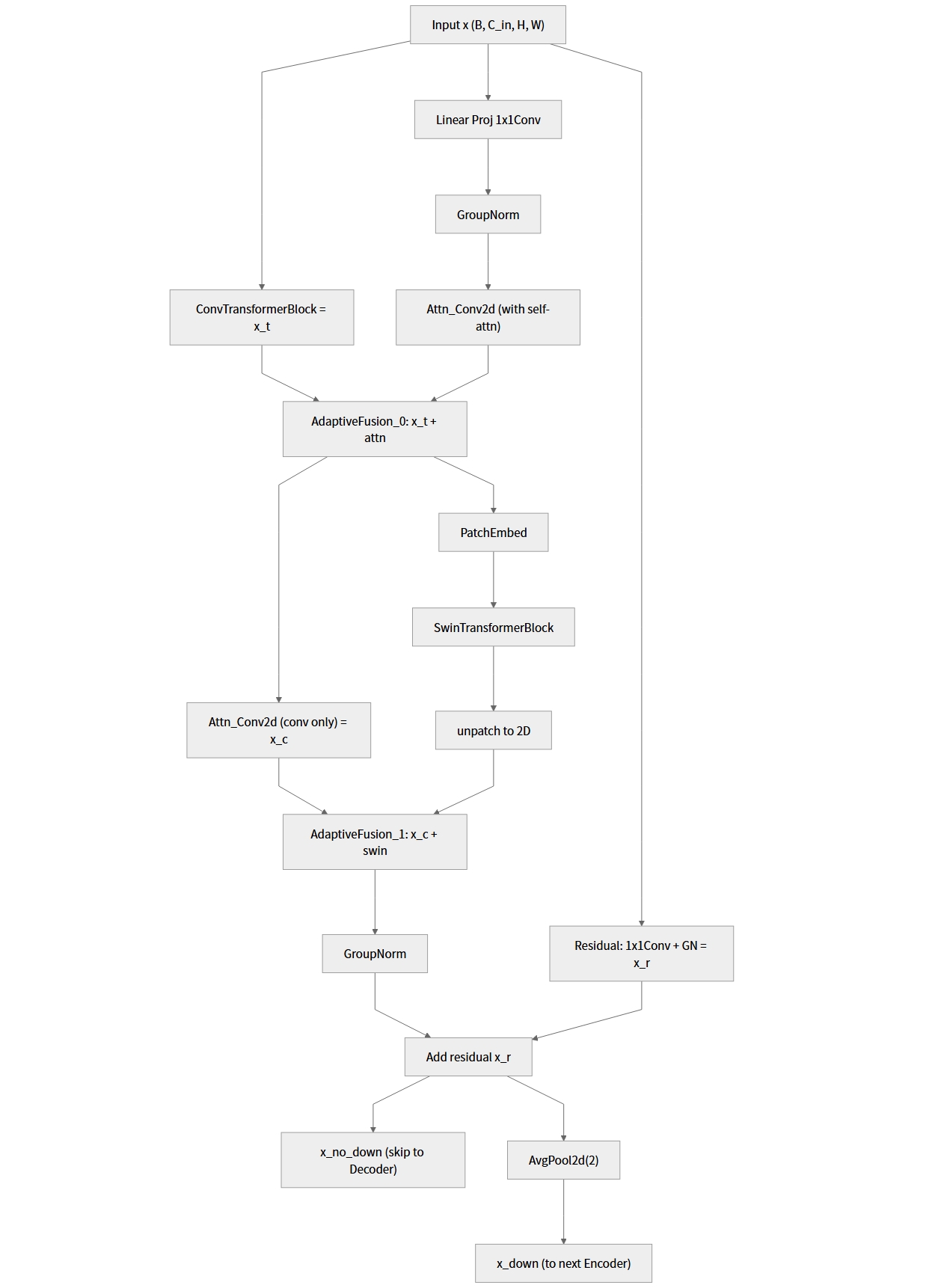}
    \caption{Downsample\_Vit: dual-pathway convolutional-transformer block. Each encoder layer employs a two-stage design: Stage~1 fuses a ConvTransformer stream with a convolutional-attention stream, and Stage~2 fuses a pure-convolution stream with a Swin Transformer stream, each via learnable AdaptiveFeatureFusion, with a residual connection and 2$\times$ downsampling.}
    \label{fig:downsample_vit}
\end{figure*}

All three branches share the same $448\times448$ input. To ensure that each branch develops distinct feature representations despite identical inputs, we adopt a \textbf{two-stage dual-pathway architecture} within each branch, illustrated in Figure~\ref{fig:downsample_vit}.

\textit{Stage 1 --- channel mixing vs.\ spatial convolution}: The branch input $x$ is processed through three parallel streams: (i)~a residual path ($1\times1$ convolution followed by GroupNorm), (ii)~a ConvTransformerBlock---a lightweight transformer that operates along the channel dimension via channel-wise MLP with LayerNorm, capturing cross-channel dependencies without the quadratic cost of spatial self-attention, and (iii)~a $1\times1$ linear projection followed by GroupNorm and an Attn\_Conv2d block (convolution with optional Local-Global Merged Attention). The ConvTransformerBlock stream and the linear-projection/Attn\_Conv2d stream are then fused via an AdaptiveFeatureFusion (AFF) module.

\textit{Stage 2 --- local convolution vs.\ windowed self-attention}: The fused features again split into two parallel streams. One stream continues through a purely convolutional Attn\_Conv2d block (attention disabled). The other stream is patch-embedded via PatchEmbed, processed by a Swin Transformer block with Window-based Multi-head Self-Attention (W-MSA), and unpatched back to the spatial domain. A second AFF module fuses these two streams. A final GroupNorm and the additive residual connection complete the block, followed by $2\times$ average-pooling downsampling.

\textbf{Why Dual-Pathway Fusion Works}:
Each encoder block (Downsample\_Vit) implements the two-stage dual-pathway design described above and illustrated in Figure~\ref{fig:downsample_vit}. The AFF fusion mechanism computes a soft weighted combination of its two input streams:
\[
\text{Fused} = \alpha \cdot \text{Stream}_1 + (1-\alpha) \cdot \text{Stream}_2,
\]
where $\alpha$ is produced by applying softmax or sigmoid to a learnable parameter, allowing the model to dynamically balance the contribution of each stream. This is not a fixed residual addition---the fusion weights are optimized during training and can vary across branches and depth levels. This two-stage design exploits the complementary strengths of the two streams. In our experiments, we observed a consistent trade-off: the convolutional stream produces globally stable but visually smooth features, while the Swin Transformer stream captures finer detail but exhibits instability at high-frequency boundaries---such as clothing edges, finger contours, and face-background transitions. The learnable AFF fusion resolves this tension: the convolutional stream acts as a stabilizing anchor that suppresses boundary jitter, while the Transformer stream injects the fine detail that pure convolution lacks. Because the fusion weights are optimized independently per branch and depth level, the stability-detail balance adapts dynamically, and the divergent discriminator-driven gradients prevent homogenization across the three parallel branches. We attribute this complementarity to the opposing inductive biases of the two operations under adversarial training: convolution acts as a low-pass filter that accumulates smoothing across layers, yielding stable but over-smoothed features, while windowed self-attention applies sharp, content-dependent weights that preserve high-frequency detail at the cost of sensitivity to boundary perturbations. Under discriminator pressure---which penalizes both blur and boundary instability---the two streams are pushed toward opposite extremes, and the learnable AFF fusion restores the balance. This learnable two-stage fusion replaces the fixed scalar blending of Swin features used in earlier iterations, where a constant weight combined the two streams. The scalar approach proved insufficient to balance the opposing stability--detail trade-off described above, motivating the fully learnable AdaptiveFeatureFusion design.

\begin{figure*}[tp]
    \centering
    \includegraphics[width=1\textwidth]{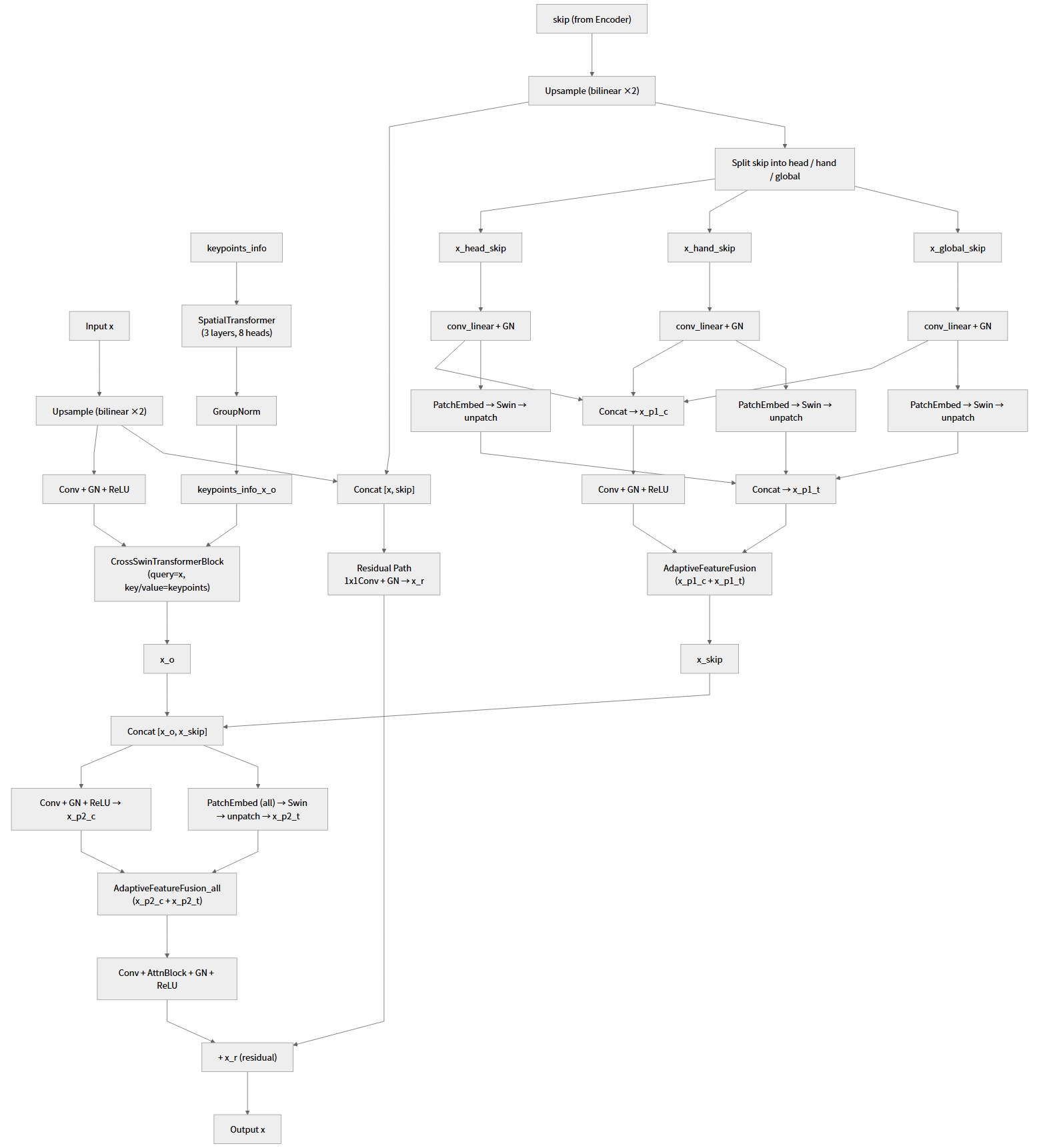}
    \caption{Upsample\_Vit: dual-pathway decoder block with skip injection.}
    \label{fig:upsample_vit}
\end{figure*}

\textbf{Decoder Upsample Blocks}:
The decoder employs a symmetric architecture (Figure~\ref{fig:img_decoder}) with Upsample\_Vit blocks (Figure~\ref{fig:upsample_vit}) that mirror the encoder's dual-pathway design. Each Upsample\_Vit block first upsamples the incoming feature map via bilinear interpolation, then concatenates the corresponding skip connection from the encoder---a concatenation of the AdaIN-fused head, hand, and global features. The combined features are split into three MoE branches (head, hand, global), each processed independently through a conv\_linear projection followed by parallel CNN and Swin Transformer pathways fused via AdaptiveFeatureFusion. At select layers (up\_3, up\_5, up\_7), keypoints\_info\_f from the MappingNetwork is injected through a SpatialTransformer and CrossSwinTransformerBlock, enabling keypoint-guided cross-attention. After MoE processing, the three branches are concatenated and passed through a second, global dual-pathway stage (PatchEmbed $\rightarrow$ Swin $\rightarrow$ AdaptiveFeatureFusion\_all), followed by convolutional refinement with Local-Global Merged Attention.

\begin{figure}[tp]
    \centering
    \includegraphics[width=0.5\textwidth]{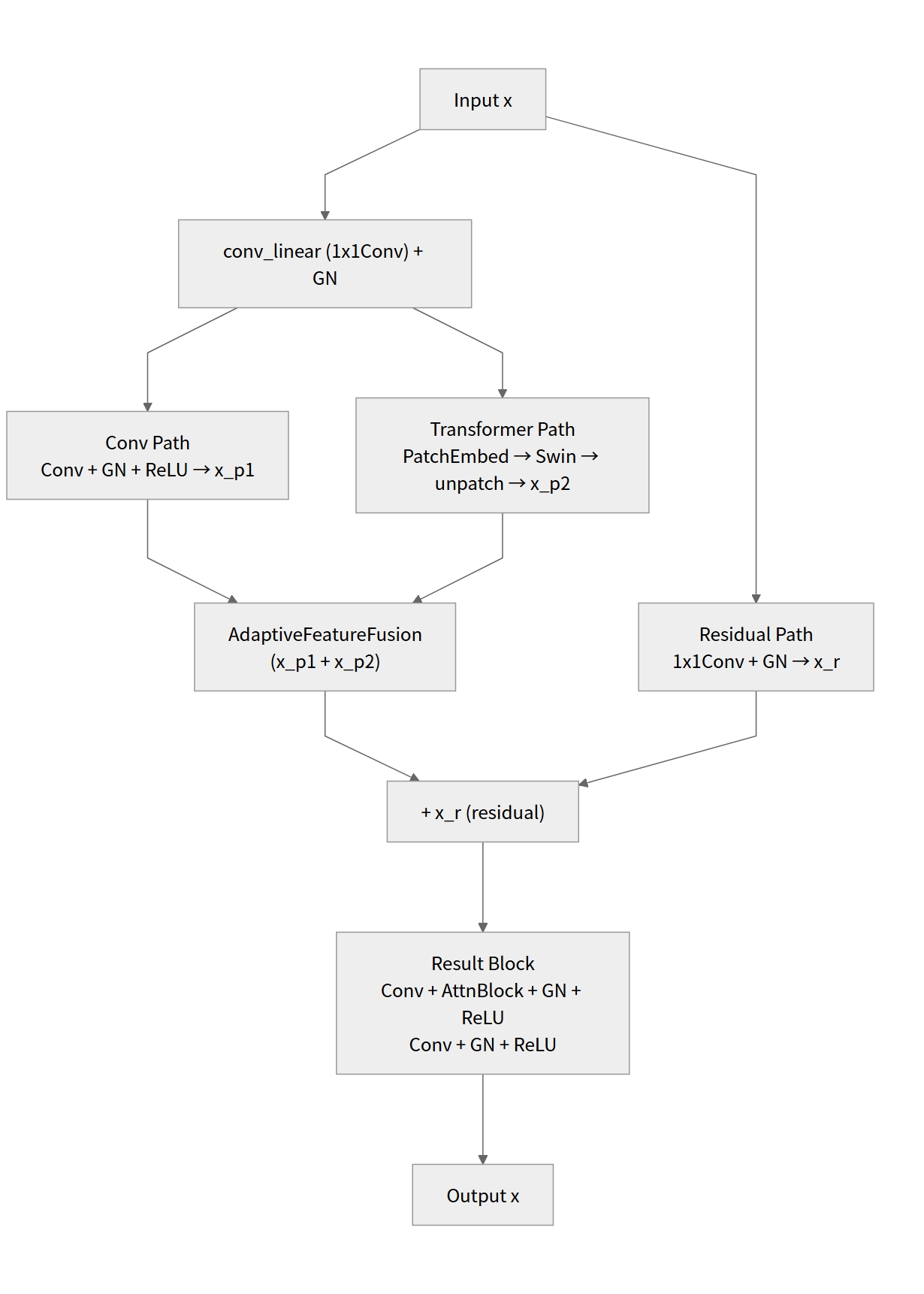}
    \caption{Upsample\_Transformer: final decoder block.}
    \label{fig:upsample_transformer}
\end{figure}

The final block, Upsample\_Transformer (Figure~\ref{fig:upsample_transformer}), is a simplified single-channel variant that omits MoE branches, skip injection, and cross-attention. It operates at the original resolution (no upsampling) and serves purely as a refinement stage: a residual path preserves the input, while the main path processes features through parallel CNN and Swin Transformer pathways fused via AdaptiveFeatureFusion, followed by two convolutional layers with intermediate attention.

\textbf{MappingNetwork}:
Keypoints are encoded by a lightweight MappingNetwork consisting of a Transformer encoder (3 layers, 8 heads) with positional embeddings. The input---133 skeleton joints with 3D coordinates---is projected to a latent dimension $z_{\text{dim}}$, processed through multi-head self-attention and MLP blocks, then flattened and linearly projected to produce keypoints\_info\_f, which is shared across all decoder layers.


\begin{figure*}[tp]
    \centering
    \includegraphics[width=0.8\textwidth]{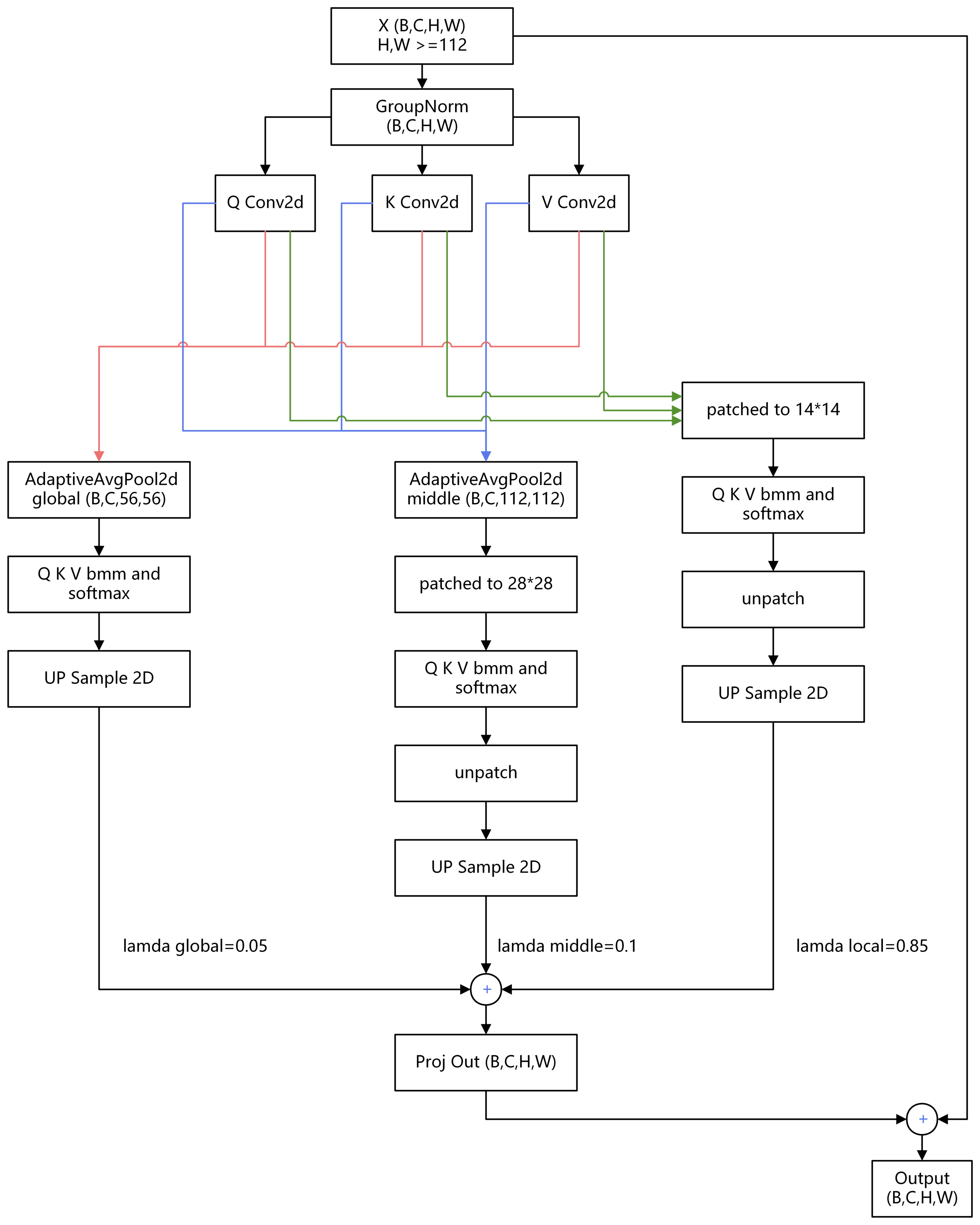}
    \caption{Local-Global Merged Attention block.}
    \label{fig:local_global_attention}
\end{figure*}

\textbf{Local-Global Merged Attention}:
We introduce a multi-scale attention mechanism that integrates local and global feature representations to improve contextual awareness (Figure~\ref{fig:local_global_attention}). The proposed approach comprises three components:

\textit{Local Attention}: Following the ViT paradigm, high-resolution feature maps are divided into 14$\times$14 patches. Query-Key-Value (QKV) computations are performed to capture local attention, and the results are reassembled to the original resolution.

\textit{Sub-Local Attention}: High-resolution feature maps are downsampled to 112$\times$112 using average pooling and partitioned into 28$\times$28 patches. After QKV computations, the patches are unpatched, and nearest-neighbor interpolation is applied to upsample the features back to the original resolution.

\textit{Global Attention}: The high-resolution feature maps are further downsampled to 56$\times$56 via average pooling. QKV computations are performed, followed by nearest-neighbor interpolation to restore the original resolution.

\begin{figure}[tp]
    \centering
    \includegraphics[width=0.2\textwidth]{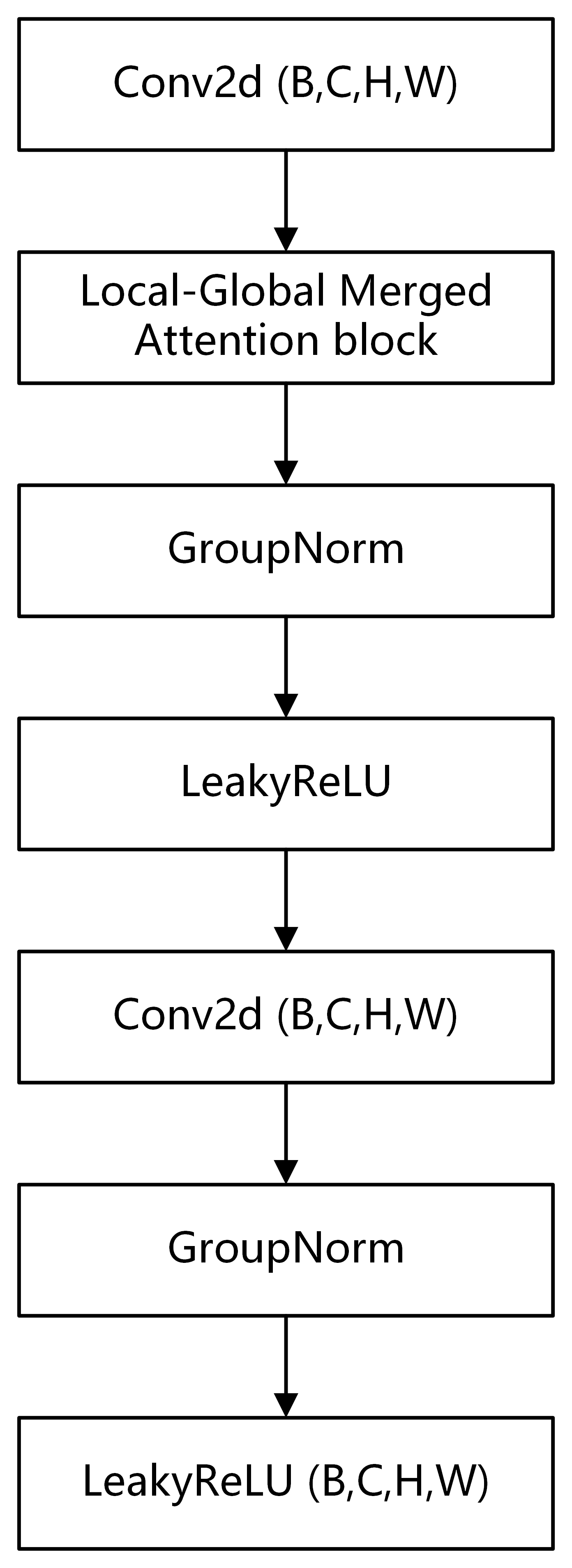}
    \caption{Conv2D with Local-Global Merged Attention block (B,C,H,W).}
    \label{fig:conv2d_local_global_merged_attention_block}
\end{figure}

\textit{Merged Attention}: The final attention map is computed as a weighted combination:
\[
\begin{aligned}
\text{Merged Attention} = 0.85 &\times \text{Local} \\
+ \; 0.1 &\times \text{Sub-Local} \\
+ \; 0.05 &\times \text{Global}.
\end{aligned}
\]
This formulation balances fine-grained local details with broader contextual information. The 0.85/0.10/0.05 weighting reflects our design principle that local detail should dominate, with sub-local and global context serving as soft constraints. This attention is instantiated as a dedicated block---Conv2D with Local-Global Merged Attention---used within Attn\_Conv2d modules (Figure~\ref{fig:conv2d_local_global_merged_attention_block}).


\textbf{AdaIN-based Style-Skeleton Fusion}:
The model takes as input two three-channel images: a skeleton image and a style image. Both are processed through separate encoders. At each encoder layer, the outputs are combined using Adaptive Instance Normalization (AdaIN). The resulting features serve as skip connections in a U-Net architecture, facilitating the integration of structural (skeleton) and stylistic (appearance) information across multiple scales. This design ensures that the generated video maintains both accurate pose structure and photorealistic texture.

\subsection{United Loss and Training Strategy}
In this work, we introduce a GAN framework that incorporates a generator and multiple discriminators. Training multiple discriminators independently poses a severe stability challenge that is most acute in the early stages of training. During this phase, gradients are large, and the three discriminators---each optimizing toward its own realism criterion---produce conflicting gradient directions. Without coordination, the generator receives incoherent feedback and frequently fails to converge, with output quality degrading progressively over successive iterations rather than improving.

Notably, we observe that this conflict is transient: as training progresses and gradients shrink, the discriminators gradually develop an implicit coordination---an emergent ``consensus'' where their individual criteria naturally align. At this later stage, the need for an explicit consensus mechanism weakens, since the discriminators have effectively learned to cooperate without external constraint. United Loss therefore functions primarily as a \emph{training wheel} that stabilizes the vulnerable early phase, while its influence naturally recedes once coordination emerges.

To provide this early-phase stability, we propose a United Loss function that unifies the training objectives of all discriminators through a soft consensus mechanism. Each discriminator computes its own adversarial loss independently, but is additionally regularized by the ensemble average of all three discriminators at a weighting of 10\% (i.e., $\lambda_{\text{united}} = 0.1$). Concretely, for each discriminator $D_i$:
\[
\mathcal{L}_{D_i}^{\text{total}} = \mathcal{L}_{D_i}^{\text{adv}} + \lambda_{\text{united}} \cdot \mathcal{L}_{\text{united}},
\]
where $\mathcal{L}_{\text{united}}$ is computed from the equally-weighted average output of all three discriminators. This formulation ensures that no single discriminator can deviate too far from the ensemble consensus during the unstable early phase, while still preserving individual specialization. The 10\% weight was chosen empirically to balance stability (higher weight) against discriminative diversity (lower weight).

The efficacy of this simple averaging mechanism can be understood through the dynamics of discriminator divergence. Without United Loss, we observe that the generator tends to \emph{capitulate} to a single discriminator---optimizing toward whichever criterion is momentarily easiest to satisfy. This creates a runaway feedback loop: the favored discriminator's loss collapses to a very low value, while the remaining discriminators' losses surge, since the generator no longer attends to their criteria. The result is de facto single-discriminator training, with the other two discriminators contributing only noise.

United Loss breaks this loop through a \emph{proportional penalty} on deviation from consensus. Since $\mathcal{L}_{\text{united}}$ is the equally-weighted average of all three losses, a discriminator whose individual loss has collapsed contributes a disproportionately large share of its own total loss through the consensus term: with $\mathcal{L}_{D_i}^{\text{adv}}$ near zero, the total loss $\mathcal{L}_{D_i}^{\text{total}} \approx \lambda_{\text{united}} \cdot \mathcal{L}_{\text{united}}$ is dominated entirely by the ensemble average, which is elevated by the other (high-loss) discriminators. Conversely, a discriminator with an already-high loss is relatively unaffected, because its total loss is dominated by its own adversarial term rather than the consensus term. The low-loss ``culprit'' is therefore penalized the most, in proportion to its deviation from the group, while the high-loss ``victims'' remain free to exert their own gradients. This asymmetry pulls the deviating discriminator back toward consensus without suppressing the legitimate specialization signals of the others.

To integrate feedback from all three discriminators, we define a weighted sum of their individual losses for the generator:
\[
\mathcal{L}_{\mathrm{united}}^{\mathrm{gen}}
= \lambda_{g}\,\mathcal{L}_{\mathrm{global}}
+ \lambda_{h}\,\mathcal{L}_{\mathrm{hand}}
+ \lambda_{f}\,\mathcal{L}_{\mathrm{head}},
\]
where $\lambda_{g}=0.33$, $\lambda_{h}=0.33$, and $\lambda_{f}=0.33$. Each loss term is computed via binary cross-entropy (BCE) on real versus generated samples.

\paragraph{Alternating Training Modes}
The training loop cycles through three fixed modes in a deterministic rotation. Let $s$ denote the training step; the active mode is determined by
\[
\text{mode}(s) = \lfloor s / 10 \rfloor \bmod 3,
\]
yielding a 1:1:1 rotation in which each mode persists for 10 consecutive steps:

\begin{enumerate}
  \item \textbf{Discriminator mode} ($\text{mode}=0$): All three discriminators are updated in parallel while the generator is frozen. If a sample contains no hands, only the global discriminator is updated; the hand and head discriminators receive no gradient on such samples.
  \item \textbf{Generator overall mode} ($\text{mode}=1$): The aggregated generator loss $\mathcal{L}_G$ is backpropagated through all generator parameters simultaneously, providing holistic guidance from the combined global and local feedback.
  \item \textbf{Generator partial mode} ($\text{mode}=2$): Each branch (head, hand, global) is updated independently using its own discriminator-specific loss and its own optimizer, enabling region-targeted optimization without cross-branch gradient conflict. For samples without hands, this mode degenerates to a global update, since the hand branch has no valid input to process.
\end{enumerate}

This deterministic rotation replaces an earlier cosine-scheduled phased-freezing design. We found empirically that cosine scheduling---whose gradual re-weighting of phases introduced an additional degree of freedom---yielded less stable convergence than the fixed 1:1:1 alternation, which imposes a rigid, predictable rhythm on the adversarial dynamics.

\paragraph{Discriminator United Loss}
For a real sample $x$ and a generated sample $\hat{x}$, let
\[
\begin{aligned}
\bigl\{&D(x),\,D(\hat{x}),\,D_{\mathrm{hand}}(x),\,D_{\mathrm{hand}}(\hat{x}), \\
&D_{\mathrm{head}}(x),\,D_{\mathrm{head}}(\hat{x})\bigr\}
\end{aligned}
\]

be the outputs of the three discriminators. After sampling real labels $r_{\mathrm{real}}\sim\mathcal{U}(0.99,1.0)$ and fake labels $r_{\mathrm{fake}}\sim\mathcal{U}(0.0,0.01)$, the combined discriminator loss incorporates the United Loss term:
\begin{equation}
  y_{\mathrm{real}}^{\mathrm{U}}
  = \frac{1}{3}\,D(x)
  + \frac{1}{3}\,D_{\mathrm{hand}}(x)
  + \frac{1}{3}\,D_{\mathrm{head}}(x)
\end{equation}
\begin{equation}
  y_{\mathrm{fake}}^{\mathrm{U}}
  = \frac{1}{3}\,D(\hat{x})
  + \frac{1}{3}\,D_{\mathrm{hand}}(\hat{x})
  + \frac{1}{3}\,D_{\mathrm{head}}(\hat{x})
\end{equation}
\begin{equation}
  \mathcal{L}_{\mathrm{disc}}^{\mathrm{U}}
  = \mathbb{E}\bigl[\ell_{\mathrm{BCE}}(y_{\mathrm{real}}^{\mathrm{U}},\,r_{\mathrm{real}})\bigr]
  + \mathbb{E}\bigl[\ell_{\mathrm{BCE}}(y_{\mathrm{fake}}^{\mathrm{U}},\,r_{\mathrm{fake}})\bigr]
\end{equation}

\paragraph{Generator United Loss}
Let $\hat{x}$ denote a generated sample. Define the fused discriminator score for the generator as
\begin{equation}
  y_{\mathrm{gen}}
  = \frac{1}{3}\,D(\hat{x})
  + \frac{1}{3}\,D_{\mathrm{hand}}(\hat{x})
  + \frac{1}{3}\,D_{\mathrm{head}}(\hat{x}).
\end{equation}
Using a randomly sampled ``real'' label $r\sim\mathcal{U}(0.99,1.0)$, the United Loss for the generator is
\begin{equation}
  \mathcal{L}_{\mathrm{gen}}^{\mathrm{U}}
  = \mathbb{E}\bigl[\ell_{\mathrm{BCE}}(y_{\mathrm{gen}},r)\bigr],
\end{equation}
where $\ell_{\mathrm{BCE}}$ denotes the binary cross-entropy loss. This equal-weight fusion prevents any single discriminator from dominating gradient direction, improving generation quality and fairness.

\section{Experiments}

\subsection{Dataset}
We train and evaluate on a custom 156\,GB dataset of sign language videos. Each video depicts a single vocabulary word, recorded with a consistent three-part structure: the signer begins in a resting pose (hands clasped in front of the abdomen), transitions into the sign language gesture, and returns to the resting pose. The resting-pose frames---which we denote as \emph{easy samples}---account for approximately 50\% of each video's duration on average, and hence roughly half of the dataset by frame count. These easy samples are highly repetitive and trivially learnable: a model that simply reproduces the resting pose achieves near-perfect reconstruction on them, inflating PSNR without reflecting the ability to generate meaningful gestures.

We therefore construct a \textbf{filtered evaluation set} that excludes all easy samples, ensuring that every reported metric targets the challenging transition and gesture frames where generation is prone to collapse or artifacts. For training, we likewise retain only a small fraction of easy samples---sufficient for the model to learn the resting pose itself, but not enough to dominate the optimization objective. Each sample consists of a skeleton image (extracted via pose estimation), a style image (for appearance transfer), and the corresponding ground-truth video frame. When evaluated on the full unfiltered evaluation set including easy samples, PSNR exceeds 36, but we consider the filtered metric far more informative for assessing generation quality on difficult poses and hand configurations.

\subsection{Training Configuration}
\begin{figure*}[tp]
    \centering
    \includegraphics[width=0.49\textwidth]{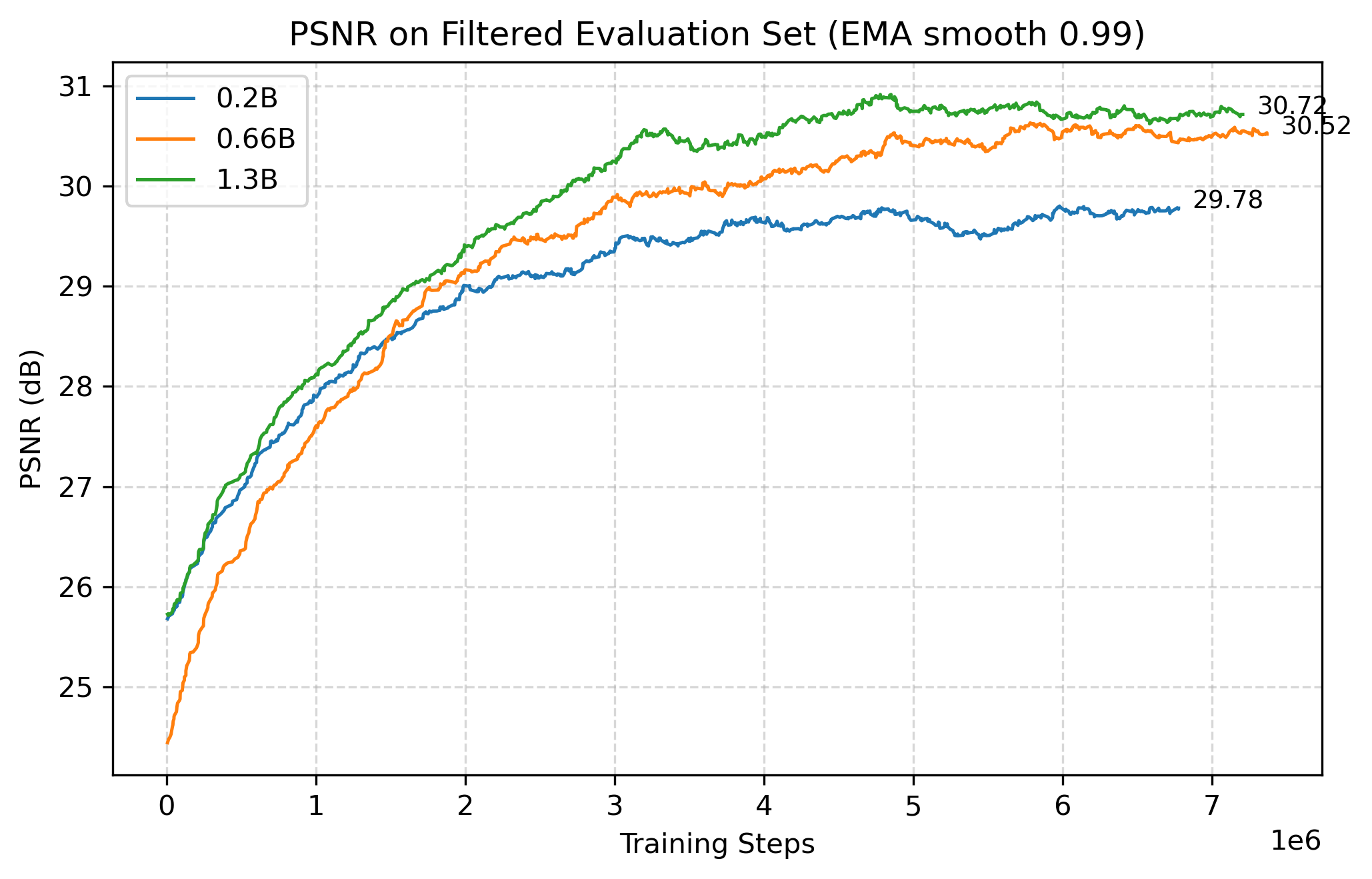}
    \hfill
    \includegraphics[width=0.49\textwidth]{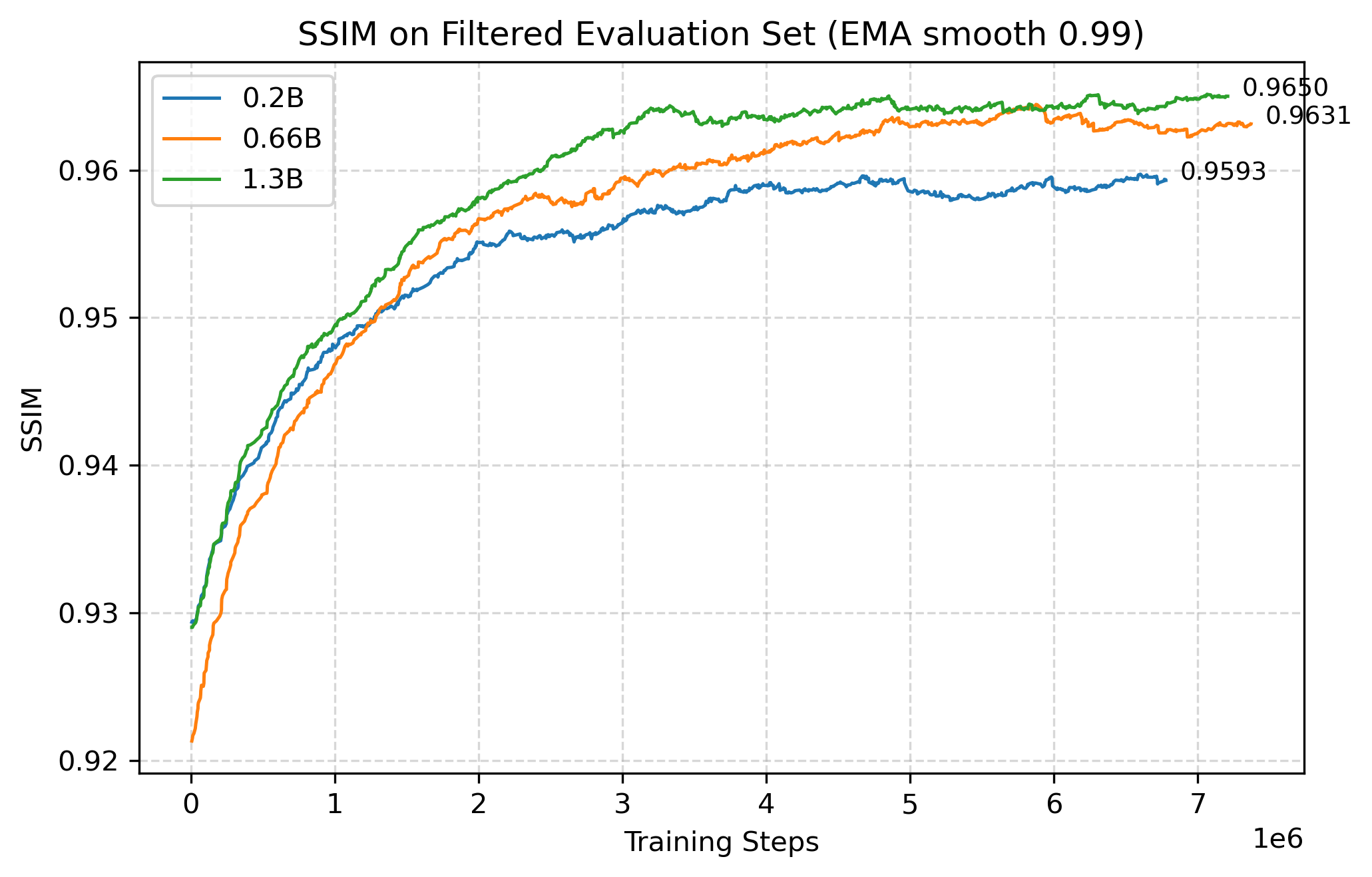}
    \caption{PSNR (left) and SSIM (right) training curves for the 0.2B, 0.66B, and 1.3B models on the filtered evaluation set, smoothed with EMA ($\alpha=0.99$). The 0.2B run is truncated at 6.78M steps because later checkpoints exhibited larger oscillation; the selected checkpoint is the most stable one. The 0.66B model lags behind 0.2B in early training and only catches up around 2M steps, while the 1.3B model consistently outperforms 0.2B from the beginning. We attribute this non-monotonic early behavior to the intermediate capacity regime of 0.66B. Final values: 0.2B = 29.78 PSNR, 0.66B = 30.52 PSNR, 1.3B = 30.72 PSNR.}
    \label{fig:psnr_ssim_curves}
\end{figure*}

All models are trained on a single consumer GPU (NVIDIA RTX 4090 48GB modified edition, functionally similar to an RTX 6000 Ada) using the AdaBelief optimizer. Training employs a cosine learning rate schedule with linear warmup over the first 50,000 steps. Full convergence requires approximately 7 million steps (2--3 months of training), with PSNR exhibiting a characteristic non-linear progression: extended plateaus punctuated by phase-transition breakthroughs, followed by terminal oscillation. We hypothesize that the plateau-to-breakthrough dynamics reflect the multi-objective optimization landscape induced by three competing discriminators with United Loss consensus---when one branch discovers a beneficial feature representation, the consensus mechanism propagates improvements to the ensemble, triggering cascading breakthroughs (Figure~\ref{fig:psnr_ssim_curves}).

\begin{figure*}[tp]
    \centering
    \includegraphics[width=0.49\textwidth]{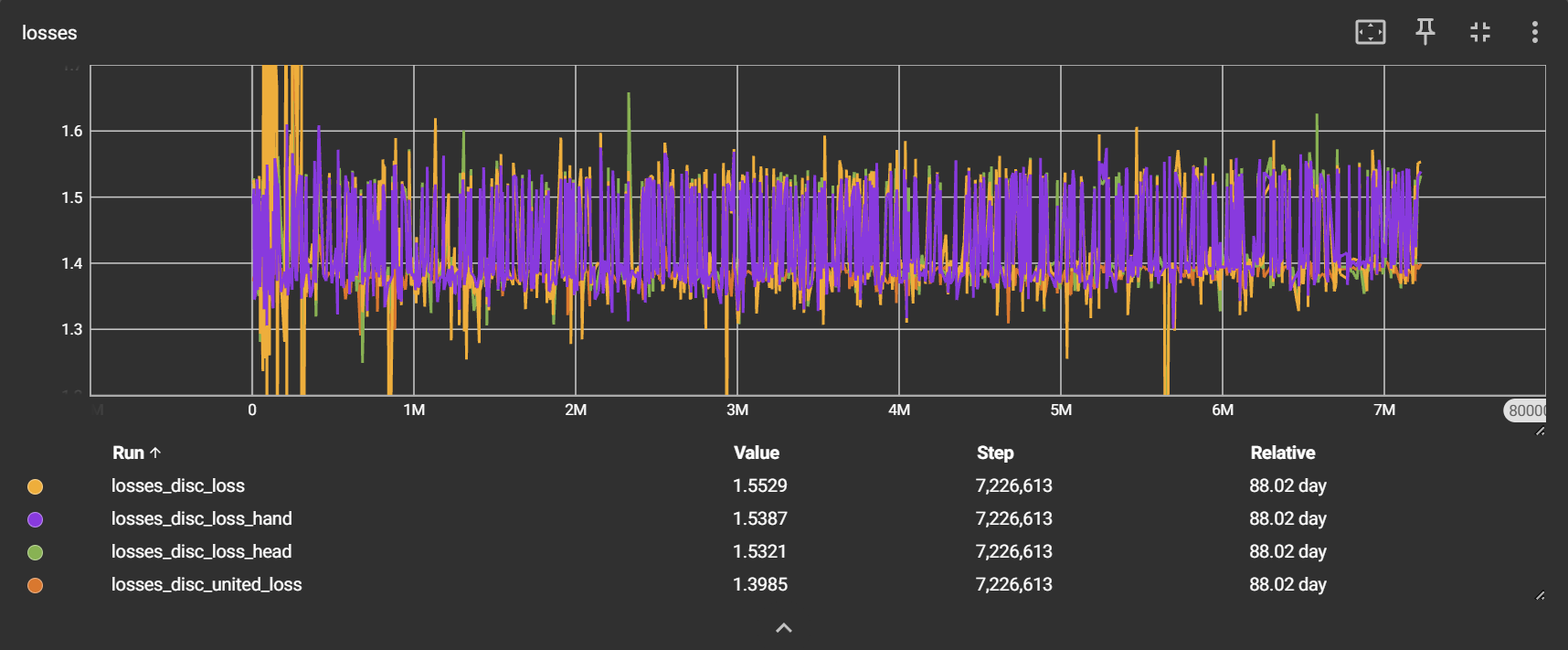}
    \hfill
    \includegraphics[width=0.49\textwidth]{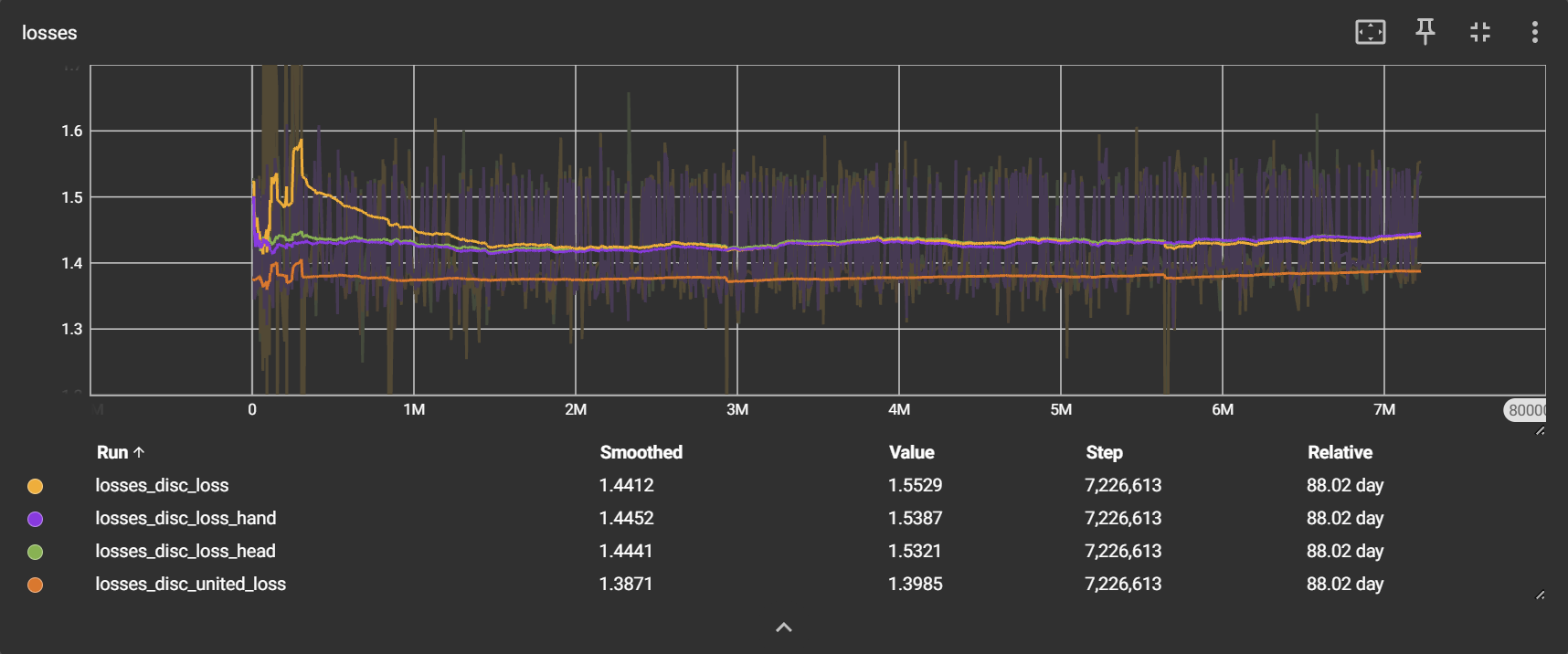}
    \caption{Discriminator losses of the 1.3B model over approximately 7.2M steps on a single RTX 4090 (48GB modified edition). Left: raw (unsmoothed) losses show persistent high-frequency adversarial fluctuations that converge to a tight band ($\approx$1.30--1.60) after the initial 0.5M steps. Right: EMA-smoothed losses show the global discriminator ($D$, yellow) spiking to $\approx$1.58 in the first 0.5M steps before descending to align with the hand ($D_{\mathrm{hand}}$, purple) and head ($D_{\mathrm{head}}$, green) losses. The United Loss term (bottom, orange) remains stable at $\approx$1.39 with no early spike, acting as a variance bound that prevents collapse. In the raw curves, the United Loss fluctuates within $\approx$1.35--1.45, a much narrower band than the individual discriminator losses, confirming its regularizing effect.}
    \label{fig:training_dynamics}
\end{figure*}

\subsection{Results: Model Scale vs.\ PSNR and SSIM}

We evaluate three model scales, all sharing the same architecture but varying in channel width:

\begin{table}[htbp]
\centering
\small
\begin{tabular}{lcccc}
\toprule
\textbf{Model} & \textbf{Params} & \textbf{PSNR} & \textbf{SSIM} & \textbf{VRAM} \\
\midrule
Small  & 0.2B  & 29.78   & 0.9593 & 1.5 GB \\
Medium & 0.66B & 30.52 & 0.9631 & $\sim$5 GB \\
Large  & 1.3B  & 30.72   & 0.9650 & 8 GB \\
\bottomrule
\end{tabular}
\caption{PSNR and SSIM across model scales, reported as smoothed peak values from the training curves (unsmoothed curves exhibit high-frequency oscillation; see Figure~\ref{fig:training_dynamics}). All values are evaluated on the filtered evaluation set (challenging cases only).}
\label{tab:psnr}
\end{table}

\begin{figure*}[t]
    \centering
    \includegraphics[width=1\textwidth]{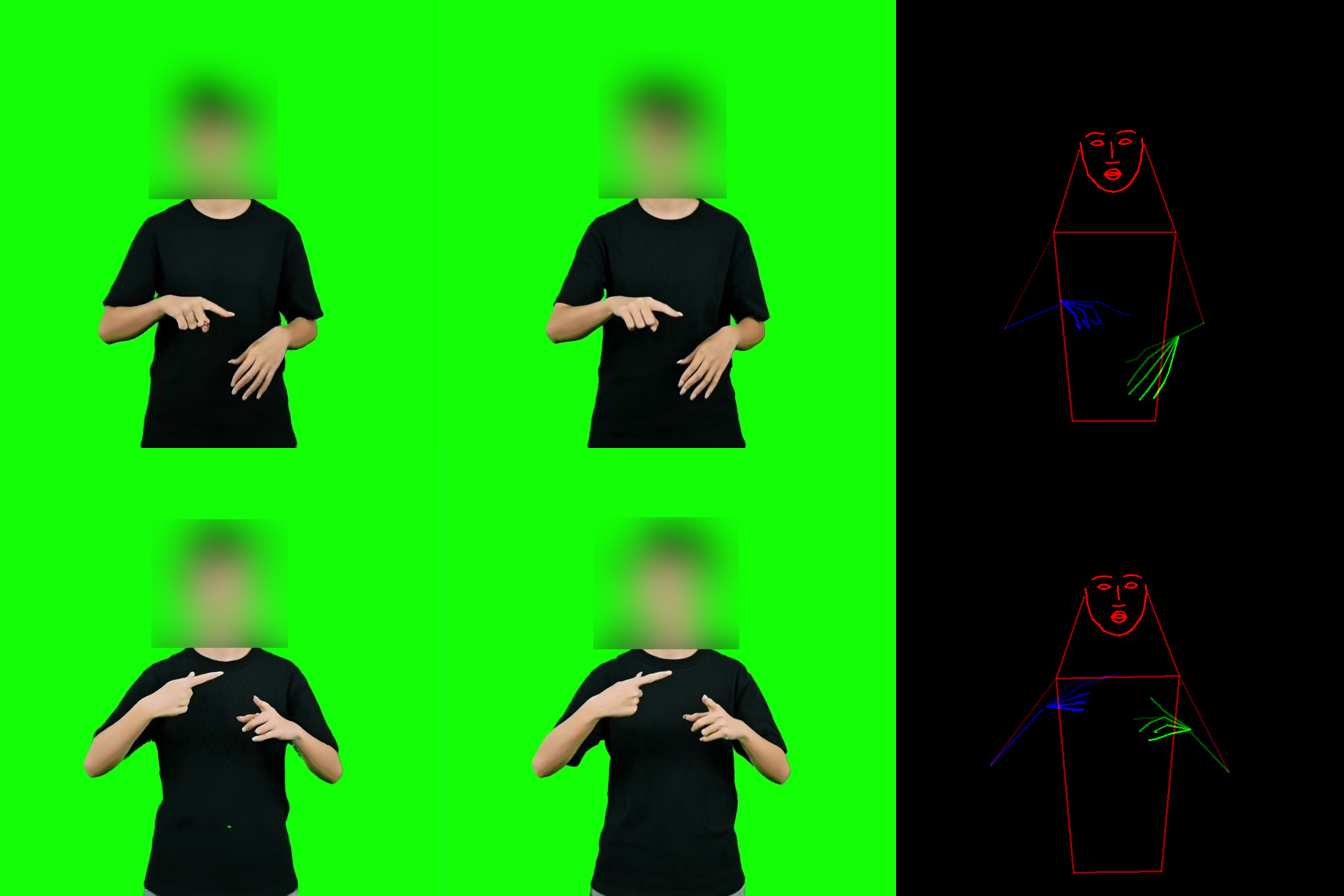}
    \caption{Qualitative results on held-out test samples from the filtered evaluation set. Each set shows (left) the frame generated by our 1.3B model, (middle) the ground truth, and (right) the driving skeleton. The two rows illustrate challenging hand configurations, including precision pinch and finger pointing. Facial regions are blurred for privacy. The model faithfully reproduces finger articulation and shading without ghosting artifacts.}
    \label{fig:qualitative_hand_samples}
\end{figure*}

The 0.66B model reaches 30.52 PSNR, capturing 79\% of the total improvement from 0.2B to 1.3B while using only half the parameters of the 1.3B variant. The gain from 0.66B to 1.3B is only +0.20 PSNR despite nearly doubling the parameter count, demonstrating sharply diminishing returns and confirming 0.66B as the sweet spot for parameter efficiency. All three variants remain deployable on consumer-grade hardware, with inference VRAM ranging from 1.5\,GB to 8\,GB.

Figure~\ref{fig:qualitative_hand_samples} provides a qualitative view on two held-out test samples with challenging hand configurations.

\subsection{Training Dynamics: Qualitative Observations}

While a controlled ablation of United Loss on the current v4 architecture has not been completed (requiring a separate 2--3 month training run), we report observational evidence consistent with the hypothesized consensus mechanism:

\begin{enumerate}
  \item \textbf{Non-linear convergence}: Training exhibits plateau-to-breakthrough phase transitions rather than smooth PSNR improvement. We observe that breakthroughs in one region (e.g., hand clarity) tend to coincide with improvements in other regions within 50,000--100,000 subsequent steps, consistent with United Loss propagating quality signals across the discriminator ensemble.
  \item \textbf{Branch specialization without collapse}: Despite all three branches receiving identical 448$\times$448 inputs, qualitative inspection of intermediate outputs confirms that the hand branch generates sharper hand regions, the face branch produces more expressive facial details, and the global branch maintains better body coherence. This implicit specialization is driven purely by discriminator-specific loss signals---the dual-pathway architecture with learnable fusion and independent optimizers prevent gradient homogenization.
  \item \textbf{Stability at scale}: Training remains stable across all three model scales without requiring gradient penalty, spectral normalization, or other common stabilization techniques beyond United Loss. We attribute this to the consensus mechanism's regularizing effect on individual discriminator updates.

\end{enumerate}

Figure~\ref{fig:training_dynamics} directly visualizes this mechanism: the United Loss term maintains a stable lower bound on the loss landscape, while individual discriminator losses converge from early-phase volatility to a coordinated regime.

\subsection{Planned Ablation: United Loss Contribution}
A controlled ablation comparing training with and without United Loss on the v4 architecture is planned but has not yet been executed due to the prohibitive training time (2--3 months per run on a single consumer GPU). The experimental design is as follows:

\begin{itemize}
  \item \textbf{Baseline (with United Loss, $\lambda_{\text{united}}=0.1$)}: Already completed; PSNR 29.78 / SSIM 0.9593 (0.2B), PSNR 30.52 / SSIM 0.9631 (0.66B), and PSNR 30.72 / SSIM 0.9650 (1.3B).
  \item \textbf{Ablation (without United Loss, $\lambda_{\text{united}}=0$)}: Planned. We hypothesize that removing United Loss would lead to (1)~increased training oscillation due to unconstrained discriminator divergence, (2)~potential branch homogenization as discriminators provide conflicting gradient signals, and (3)~reduced final PSNR, consistent with observations from an earlier ablation study conducted during the early development of the v4 architecture, where United Loss demonstrated a measurable stabilizing effect.
  \item \textbf{Varied $\lambda_{\text{united}}$}: Planned. Exploring the trade-off between consensus strength (higher $\lambda$) and discriminative diversity (lower $\lambda$).
\end{itemize}

\section{Conclusion}

\subsection{Summary of Contributions}
We have presented a loss-guided multi-expert GAN framework for sign language video synthesis featuring: (1)~three specialized discriminators (global, hand, face) with Haar wavelet preprocessing; (2)~a Multi parallel U-Net generator where three independent encoder-decoder branches specialize on distinct visual regions, driven by corresponding discriminators; (3)~a United Loss consensus mechanism that stabilizes multi-discriminator training through a 10\% soft blending of individual losses with the ensemble average; (4)~a dual-pathway convolutional-transformer architecture with AdaptiveFeatureFusion---a learnable blending mechanism that fuses parallel convolutional and Swin Transformer streams at each encoder block, ensuring feature divergence across branches; (5)~a Local-Global Merged Attention mechanism integrating three spatial scales; and (6)~an alternating three-mode training schedule (discriminator / holistic / branch-specialized) with independent per-branch optimizers. On a filtered evaluation set targeting challenging cases, our architecture achieves 29.78--30.72 PSNR and 0.9593--0.9650 SSIM across model scales, with inference deployable on consumer-grade hardware.

\subsection{Limitations}
This work has several limitations that we aim to address in future iterations:

\textbf{Training Cost:} Full training requires 2--3 months on a single consumer GPU, which has precluded comprehensive ablation studies of individual architectural components (United Loss, dual-pathway architecture, alternating training modes, multi-scale attention). Each ablation would require a full retraining cycle, making systematic component-wise analysis infeasible under current resource constraints.

\textbf{Ablation Gap:} The contribution of United Loss to training stability and final PSNR has not been isolated on the current v4 architecture. While observational evidence---non-linear convergence dynamics, maintained branch specialization, and training stability without external regularization---is consistent with the hypothesized consensus mechanism, rigorous causal attribution requires controlled experiments.

\textbf{Scaling Bottleneck:} The current architecture activates all three branches at both training and inference time. Extending to more specialized branches (e.g., separate left-hand and right-hand experts, facial sub-region experts, or multiple domain-specific experts) would incur linear cost increases in both compute and VRAM. This scalability limitation directly motivates our ongoing work on label-routed expert selection, where a pre-trained router activates only the relevant expert branches before generation begins, maintaining bounded inference cost regardless of the number of experts.

\textbf{Metric Limitations:} We currently report PSNR and SSIM. Comprehensive evaluation with perceptual metrics (FID, LPIPS), action accuracy, and human evaluation is planned but not yet completed.

\subsection{Version History}
The current architecture (v4) is the result of iterative development over approximately two years. We document the progression below. Note that model and dataset versions evolved independently; the dataset also progressed through v1--v4, with v4 (current) increasing image contrast for improved visual appeal at a cost of approximately 0.1--0.2 PSNR relative to v3. PSNR was rigorously recorded only on v4; earlier versions were evaluated subjectively due to resource constraints.

\begin{itemize}
  \item \textbf{v1 --- AdaIN as Style Transfer}: The initial approach concatenated the skeleton (content) and style images along the channel dimension before feeding them to a standard U-Net. This caused information entanglement in deep encoder layers, where skeleton structure and style appearance became indistinguishable; the decoder could not reliably reconstruct fine details, frequently producing hands with incorrect numbers of fingers. Using two separate encoders or concatenating outputs at every layer would preserve information but double the decoder's channel count, making training infeasible under consumer GPU memory constraints. The key observation was that sign language generation could be reformulated as a \textit{style transfer} problem: the skeleton defines spatial structure, while the style image provides appearance. Replacing concatenation with Adaptive Instance Normalization (AdaIN) at each encoder layer---processing content and style through two encoder streams that remain independent until AdaIN fusion---preserved both structural and stylistic information throughout the downsampling hierarchy without increasing decoder complexity. This dual-stream design stabilized finger generation and formed the foundation for all subsequent versions.

  \item \textbf{v2 --- Keypoint-Guided Generalization}: Although v1 achieved stable hand generation, a generalization problem emerged. Skeleton sequences between vocabulary items were algorithmically interpolated, and coordinate normalization was necessary to prevent the person from drifting across the frame during transitions. However, this normalization introduced a train-test mismatch: the same gesture could appear at different normalized positions during training versus inference, occasionally causing the model to break down. Applying dropout to simulate coordinate noise---a standard regularization technique---produced negligible improvement, suggesting the deficiency was not in image-level robustness but in the model's \textit{awareness of skeleton coordinates}. Inspired by StyleGAN3's use of latent codes for fine-grained control, we embedded the 133-joint 3D keypoint coordinates into the decoder via a lightweight MappingNetwork (a 3-layer Transformer encoder) and injected the resulting keypoints\_info\_f at select decoder layers. This gave the model explicit coordinate knowledge as a form of ``informative noise,'' dramatically improving generalization: as long as coordinate shifts were not extreme, generation remained stable.

    \item \textbf{v3 --- Local Hand Discriminator}: With generalization resolved, the system reached a commercially viable quality level. Deaf users found the output intelligible and satisfactory. However, fine details remained unconvincing: fingers appeared unnaturally smooth, lacking shadows and texture. We hypothesized that the global discriminator's 448$\times$448 receptive field could not provide sufficiently granular gradient signals for small hand regions. Exploiting the skeleton coordinates already available, we cropped a 112$\times$112 region centered on each hand and trained an additional, lightweight hand discriminator operating on these patches. The generator loss was extended to $\mathcal{L}_{\text{main}} + 0.25 \cdot \mathcal{L}_{\text{hand}}$, with the two discriminators trained fully independently. Notably, while the two discriminators exhibited mild oscillation during training, this drift remained controllable within an overall convergent trajectory. Because convergence was ultimately achieved, the potential severity of multi-discriminator dynamics was not yet apparent at this early exploratory stage---the pronounced instability that would later motivate United Loss only emerged when a third discriminator was introduced in v4.

  \item \textbf{v4 (current) --- Multi-Expert Parallel Architecture}: With the company's immediate commercial risks resolved, we turned to a more fundamental question: on a consumer GPU (RTX 4090), was the model allocating a substantial fraction of its capacity to learning unimportant details such as clothing wrinkles and folds? We hypothesized that partitioning the generator into specialized expert branches---each guided by a dedicated discriminator toward a distinct visual region---would concentrate representational capacity on perceptually critical areas (hands, face) while maintaining global coherence.

  However, adding the third (head) discriminator immediately destabilized training. Whereas the two-discriminator system of v3---despite mild oscillation---remained within a controllable, convergent trajectory, the three-discriminator system exhibited dynamics reminiscent of the \emph{three-body problem}: without an external consensus constraint, the three optimizers drifted into chaotic, non-convergent trajectories. This instability directly motivated the United Loss consensus mechanism described in Section~3.3, which acts as the external constraint that binds the three discriminators into a stable configuration. The resulting architecture---three encoder-decoder branches with identical 448$\times$448 input, a dual-pathway convolutional-transformer design with AdaptiveFeatureFusion, United Loss consensus across three discriminators, and an alternating three-mode training schedule---is the first version for which we report systematic PSNR and SSIM measurements: 29.78 / 0.9593 (0.2B), 30.52 / 0.9631 (0.66B), and 30.72 / 0.9650 (1.3B) on the filtered evaluation set of dataset v4.
\end{itemize}

\subsection{Future Work}
Beyond completing the planned ablation studies, our primary research direction is extending the loss-guided expert specialization paradigm---validated here in the GAN framework---to diffusion models. Our ongoing proposal, \textit{Loss-Guided MoE Diffusion with Label-Routed Experts}, aims to replace full branch activation at inference with a pre-trained label router that selects only relevant expert branches before generation begins. This would preserve the quality gains of multi-expert architectures while keeping inference cost bounded, directly addressing the scaling bottleneck identified in the current work.

\subsection{Acknowledgments}
The system was showcased at the 2025 Hong Kong Frontier Technology Summit. We thank the event organizers and attendees for valuable feedback.

\bibliography{references}

\clearpage

\appendix
\section{Training Algorithms (Pseudocode)}

\RaggedRight
\sloppy
\small
\begin{algorithm*}
\caption{Main Training Loop}
\begin{algorithmic}[1]
\State \textbf{Initialize:} 
\State $G, D, D_{hand}, D_{head}$ \Comment{Generator and Discriminators}
\State $opt_G, opt_D, opt_{D_{hand}}, opt_{D_{head}}$ \Comment{Optimizers}
\State Hyperparameters: $\lambda_{hand}, \lambda_{head}, \lambda_{L1}, \lambda_{united}$

\For{each training step $t$}
    \State $(content, target, style) \gets \text{LoadBatch}()$
    \State $(bodies, faces, left\_hands, right\_hands) \gets \text{ProcessKeypoints}(npy\_list)$
    \State $(hand\_input, real\_hands, head\_input, real\_head) \gets \text{CropRegions}(content, target)$
    
    \State $mode \gets (\lfloor t / 10 \rfloor) \bmod 3$ \Comment{Fixed 1:1:1 rotation}
    
    \If{$mode == 0$}
    \State $\text{TrainDiscriminators}(...)$
\ElsIf{$mode == 1$}
    \State $\text{TrainGenerator}(..., \text{overall})$
\Else
    \State $\text{TrainGenerator}(..., \text{partial})$
\EndIf
    
    \If{$t \mod 20000 == 0$}
        \State $\text{UpdateLearningRates}()$
    \EndIf
    
    \If{$t \mod \text{val\_gap} == 0$}
        \State $\text{RunValidation}()$
    \EndIf
    
    \If{$t \mod \text{save\_gap} == 0$}
        \State $\text{SaveCheckpoints}()$
    \EndIf
\EndFor
\end{algorithmic}
\end{algorithm*}

\begin{algorithm*}
\caption{TrainGenerator Procedure}
\begin{algorithmic}[1]
\Procedure{TrainGenerator}{$ $}
    \State $gen\_output \gets G(content, style, bodies, faces, left\_hands, right\_hands)$
    \State $(gen\_hands, gen\_head) \gets \text{CropGeneratedRegions}(gen\_output)$
    
    \State \Comment{With discriminators frozen}
    \State $D\_fake \gets D(content, gen\_output)$
    \State $D\_hand\_fake \gets D_{hand}(hand\_input, gen\_hands)$
    \State $D\_head\_fake \gets D_{head}(head\_input, gen\_head)$
    
    \State $L_{GAN} \gets BCE(D\_fake, 1)$
    \State $L_{GAN\_hand} \gets \lambda_{hand} \cdot BCE(D\_hand\_fake, 1)$
    \State $L_{GAN\_head} \gets \lambda_{head} \cdot BCE(D\_head\_fake, 1)$
    \State $L_{L1} \gets \lambda_{L1} \cdot \|target - gen\_output\|_1$
    \State $L_{united} \gets BCE(\text{avg}(D\_fake, D\_hand\_fake, D\_head\_fake), 1)$
    \State $L_G \gets \sum \text{losses}$
    
    \State $opt_G.\text{zero\_grad}()$
    \State $opt_{Global}.\text{zero\_grad}()$
    \State $opt_{Hand}.\text{zero\_grad}()$
    \State $opt_{Head}.\text{zero\_grad}()$

    \State $L_{Global} \gets L_{GAN} + \lambda_{Gu} \cdot L_{united} + L_{L1}$
    \State $L_{hand} \gets L_{GAN\_hand} + \lambda_{Gu} \cdot L_{united} + L_{L1}$
    \State $L_{head} \gets L_{GAN\_head} + \lambda_{Gu} \cdot L_{united} + L_{L1}$
    
    \State \Comment{If mode == 1 (overall):}
\State $L_G.\text{backward}()$
\State $opt_G.\text{step}()$
\State \Comment{If mode == 2 (partial):}
\State $L_{Global}.\text{backward}()$
\State $opt_{Global}.\text{step}()$
\State $L_{hand}.\text{backward}()$
\State $opt_{Hand}.\text{step}()$
\State $L_{head}.\text{backward}()$
\State $opt_{Head}.\text{step}()$
\EndProcedure
\end{algorithmic}
\end{algorithm*}

\begin{algorithm*}
\caption{TrainDiscriminators Procedure}
\begin{algorithmic}[1]
\Procedure{TrainDiscriminators}{$ $}
    \State \Comment{With generator frozen}
    \State $gen\_output \gets G(content, style, bodies, faces, left\_hands, right\_hands)$
    \State $(gen\_hands, gen\_head) \gets \text{CropGeneratedRegions}(gen\_output)$
    
    \State $D\_real \gets D(content, target)$
    \State $D\_fake \gets D(content, gen\_output)$
    \State $D\_hand\_real \gets D_{hand}(hand\_input, real\_hands)$
    \State $D\_hand\_fake \gets D_{hand}(hand\_input, gen\_hands)$
    \State $D\_head\_real \gets D_{head}(head\_input, real\_head)$
    \State $D\_head\_fake \gets D_{head}(head\_input, gen\_head)$
    \State $L_{united} \gets BCE(\text{avg}(D\_real, D\_hand\_real, D\_head\_real), 1) + BCE(\text{avg}(D\_fake, D\_hand\_fake, D\_head\_fake), 0)$
    \State $L_D \gets BCE(D\_real, 1) + BCE(D\_fake, 0) + \lambda_{Du} \cdot L_{united}$
    \State $L_{D\_hand} \gets BCE(D\_hand\_real, 1) + BCE(D\_hand\_fake, 0) + \lambda_{Du} \cdot L_{united}$
    \State $L_{D\_head} \gets BCE(D\_head\_real, 1) + BCE(D\_head\_fake, 0) + \lambda_{Du} \cdot L_{united}$
    
    \State $opt_D.\text{zero\_grad}()$
    \State $opt_{D_{hand}}.\text{zero\_grad}()$
    \State $opt_{D_{head}}.\text{zero\_grad}()$
    \State $L_D.\text{backward}()$
    \State $L_{D\_hand}.\text{backward}()$
    \State $L_{D\_head}.\text{backward}()$
    \State $\text{UpdateAllDiscriminators}()$
\EndProcedure
\end{algorithmic}
\end{algorithm*}

\begin{algorithm*}
\caption{Helper Procedures}
\begin{algorithmic}[1]
\Procedure{ProcessKeypoints}{$npy\_list$}
    \State $bodies \gets npy\_list[:, 0:23, :]$
    \State $faces \gets npy\_list[:, 23:91, :]$
    \State $left\_hands \gets npy\_list[:, 91:112, :]$
    \State $right\_hands \gets npy\_list[:, 112:, :]$
    \State \Return $(bodies, faces, left\_hands, right\_hands)$
\EndProcedure

\Procedure{CropRegions}{$content, target$}
    \State Detect hand and head regions from keypoints
    \State $hand\_input \gets \text{CropHandInput}(content)$
    \State $real\_hands \gets \text{CropRealHands}(target)$
    \State $head\_input \gets \text{CropHeadInput}(content)$
    \State $real\_head \gets \text{CropRealHead}(target)$
    \State \Return $(hand\_input, real\_hands, head\_input, real\_head)$
\EndProcedure

\Procedure{UpdateLearningRates}{}
    \State $G.\text{update\_lr}()$
    \State $D.\text{update\_lr}()$
    \State $D_{hand}.\text{update\_lr}()$
    \State $D_{head}.\text{update\_lr}()$
    \State Update part-specific learning rates
\EndProcedure
\end{algorithmic}
\end{algorithm*}

\begin{algorithm*}
\caption{Validation and Utility Procedures}
\begin{algorithmic}[1]
\Procedure{RunValidation}{}
    \State Load validation batch
    \State Generate predictions
    \State Calculate metrics
    \State Save sample images
    \State Log to TensorBoard
\EndProcedure

\Procedure{SaveCheckpoints}{}
    \State Save generator weights
    \State Save discriminators weights
    \State Save optimizer states
\EndProcedure

\Procedure{UpdatePartParameters}{}
    \For{each part (head, hand, global)}
        \State Calculate part-specific loss
        \State Backpropagate through part
        \State Update part parameters
    \EndFor
\EndProcedure
\end{algorithmic}
\end{algorithm*}

\subsection*{Notation}
\begin{itemize}
\item $G$: Generator, $D$: Main (global) discriminator
\item $D_{hand}$: Hand discriminator, $D_{head}$: Head discriminator
\item $\lambda$: Loss weights, BCE: Binary cross-entropy
\item avg: Equally-weighted average of discriminator outputs
\end{itemize}

\subsection*{Key Features}
\begin{itemize}
\item Alternating training modes (discriminator / generator overall / generator partial) with fixed 1:1:1 rotation

\item Multi-scale discriminators (global, hand, head) with United Loss consensus
\item Part-specific optimization with independent optimizers per generator branch
\item Combined adversarial and L1 losses for generator
\item Cosine learning rate scheduling with linear warmup
\item Regular validation and checkpointing
\end{itemize}

\end{document}